\documentclass{article}
\usepackage{spconf,amsmath,graphicx}
\usepackage{amssymb}
\usepackage{arydshln}

\usepackage{times}
\usepackage{epsfig}
\usepackage{caption}
\usepackage{algorithm}
\usepackage{bbm}
\usepackage{bm}
\usepackage{color}
\usepackage{multirow}
\usepackage{booktabs}
\usepackage{diagbox}
\usepackage[table]{xcolor}
\usepackage[pagebackref,breaklinks,colorlinks,linkcolor=black,anchorcolor=black,citecolor=black]{hyperref}
\usepackage{soul}

\newcommand{\etal}{\textit{et al}. }
\newcommand{\ie}{\textit{i}.\textit{e}.}
\newcommand{\eg}{\textit{e}.\textit{g}.}

\newcommand{\Tref}[1]{Table~\ref{#1}}

\newcommand{\Fref}[1]{Figure~\ref{#1}}

\newcommand{\eref}[1]{Eq.~(\ref{#1})}
\newcommand{\fref}[1]{Fig.~\ref{#1}}

\title{Hierarchical Terrain Attention and Multi-Scale Rainfall Guidance For Flood Image Prediction}
\name{Feifei Wang\textsuperscript{\rm 1} \qquad Yong Wang\textsuperscript{\rm 1} \qquad  Bing Li\textsuperscript{\rm 2}\qquad Qidong Huang\textsuperscript{\rm 1}\qquad Shaoqing Chen\textsuperscript{\rm 1,$\ast$}\thanks{$\ast$ Corresponding author. Email: windcsq@ustc.edu.cn.}}

\address{$^{1}$ University of Science and Technology of China \\
$^{2}$ China Water Sunny Data Technology Co., Ltd.}

\begin{document}
%
\maketitle
\begin{abstract}

With the deterioration of climate, the phenomenon of rain-induced flooding has become frequent. To mitigate its impact, recent works adopt convolutional neural network or its variants to predict the floods. However, these methods directly force the model to reconstruct the raw pixels of flood images through a global constraint, overlooking the underlying information contained in terrain features and rainfall patterns. To address this, we present a novel framework for precise flood map prediction, which incorporates hierarchical terrain spatial attention to help the model focus on spatially-salient areas of terrain features and constructs multi-scale rainfall embedding to extensively integrate rainfall pattern information into generation. To better adapt the model in various rainfall conditions, we leverage a rainfall regression loss for both the generator and the discriminator as additional supervision. Extensive evaluations on real catchment datasets demonstrate the superior performance of our method, which greatly surpasses the previous arts under different rainfall conditions.

\end{abstract}
\begin{keywords}
Flood prediction, image translation, GAN
\end{keywords}

\section{Introduction}
\label{sec:intro}

As one of the most common disasters, rain-induced floods has drawn increasing attention in recent years \cite{8802923,fernandes2022river}. 
Traditional countermeasures against flooding mainly rely on real-time river monitoring, which requires sensitive equipment like high-precision sensors. 
But unfortunately, such equipment is usually too expensive that it is difficult to be available in many places. 
Therefore, flood prediction has become a better alternative that can timely forecast the maximum water depth with little cost through modeling the evolution of floods. 
To investigate flood modeling, a variety of methods have been proposed, where deep learning based solutions have been recently shown to outperform time-consuming hydrodynamic simulation \cite{bentivoglio2021deep,8451011,bazartseren2003short,kabir2020deep}. 
For instance, Guo \etal \cite{Guo2020DatadrivenFE} and Löwe \etal \cite{LOWE2021126898} used convolutional neural network to predict water depth and directly injected rainfall pattern vector into prediction through simple feature fusion.
Qian \etal \cite{qian2019physics} utilized conditional generative adversarial networks \cite{MirzaO14} to predict the next state of the flood evolution based on the current state. 
These methods, however, can not take the sufficient advantage of the given terrain or rainfall information.
\textbf{First}, they apply the global low-level constraint to reconstruct flood images during training, overlooking the spatial discrepancy of input terrain characteristics. 
It is not reasonable since terrain images are spatially-differentiated and the areas with significant changes need to be paid more attention, \eg, the terrain of steep areas deserves more concern than that of flat areas. 
\textbf{Second}, they just use the simple fusion to integrate the rainfall information into prediction instead of fully exploiting it.
With the absence of high-level guidance contained in terrain features and rainfall patterns, the learned prediction model shows really weak generalization ability on those unseen rainfall conditions.

To tackle with these limitations, we propose a novel framework for flood image prediction. 
The basic objective is to learn a generator $G$ that can translate terrain images into high-fidelity flood images (\ie, water depth maps) when conditioned by a particular rainfall pattern, similar with image translation \cite{ZhuPIE17,wang2018high, ZhanYWZLZ22,IsolaZZE17}. 
To sufficiently utilize terrain features, we introduce hierarchical terrain spatial attention (HTA) to encoder-decoder shortcuts during generation, where the attention map provides the weight penalty for the areas that have larger importance. 
This allows $G$ to focus on the spatially-salient areas at different scales that are more prone to water depth changes, facilitating the optimization of water depth reconstruction. 
To extensively explore rainfall effects, we propose multi-scale rainfall embedding (MRE) where a rainfall mapping network is trained along with $G$ from scratch.
The mapping network learns to extract rainfall feature maps in multiple scales and integrate these features into different stages of water depth map generation. 
Like vanilla GAN supervision \cite{GoodfellowPMXWOCB14,huang2021initiative}, $G$ delves into predicting water depth maps that can fool the the discriminator $D$ and $D$ needs to distinguish between the generate water depth map and the ground-truth counterpart. 
The difference is that we add a rainfall regression loss to the training objective of both $G$ and $D$, encouraging them to regress the corresponding rainfall pattern.
Note that we do not require the regressed rainfall pattern to be entirely correct, since we just use the regression as extra supervision to help the model comprehend the high-level information hidden in flood images. 


Our main contributions are:
\begin{itemize}
    \item We present a novel framework for flood image prediction, which integrates vanilla adversarial supervision with additional rainfall pattern regression to better adapt to multiple flood conditions.
    
    \item To boost the generalization ability on different catchments and rainfall conditions, we propose feature fusion structure with densely embedded information. The model is encouraged to fully exploit the terrain effects through hierarchical spatial attention and precisely identity the given rainfall information based on multi-scale rainfall embedding.
    
    \item Extensive evaluations on two catchments prove the effectiveness of the proposed framework, which greatly outperforms previous methods.

\end{itemize}

\section{Method}

\begin{figure*}[t]
\centering
\vspace{-1em}
\includegraphics[width=0.9\linewidth]{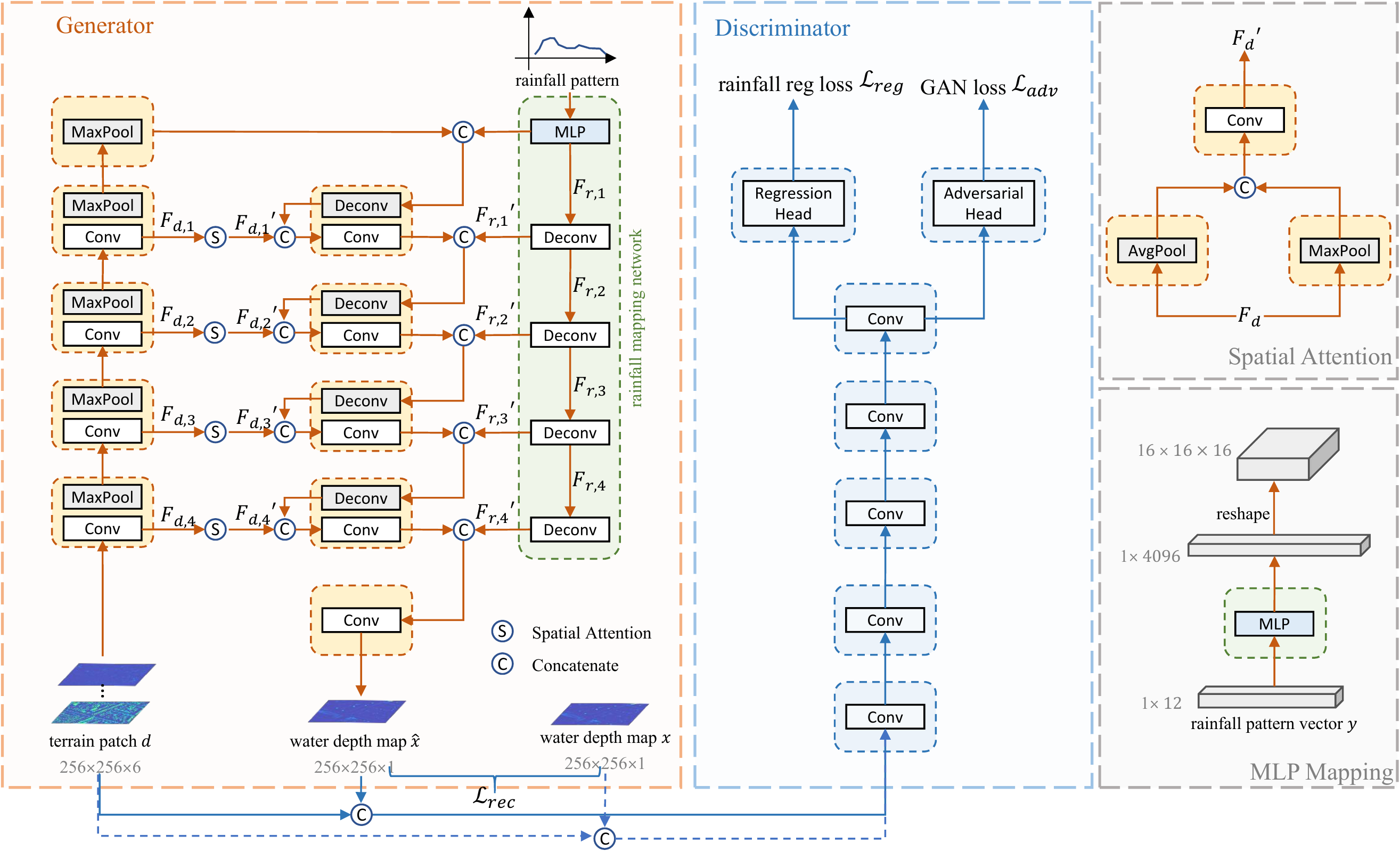}
\vspace{-1em}
\caption{The pipeline of the proposed framework. During encoding, the generator $G$ not only  aggregate features with different terrain attention maps, but also the multi-scale rainfall embedding. The discriminator $D$ learns to figure out whether the input is real or fake, while inferring the rainfall pattern of the real images to improve its discriminative ability.}
\label{fig:method}
\end{figure*}

This section gives a detailed illustration for the proposed framework, including hierarchical terrain spatial attention, multi-scale rainfall embedding and loss functions. 
Our basic objective is to predict water depth maps with the given 6-channel terrain image that depicts geomorphic features and 12-dimensional rainfall pattern vector that reflects rainfall conditions. The
overall pipeline is shown in \fref{fig:method}.


\subsection{Flood Prediction Framework}

Since the terrain images are often in large sizes, we randomly cut it into small patches  $d\in\mathbb{R}^{256\times 256\times6}$ for efficiency. 
Each terrain patch is translated into its corresponding water depth patch $\hat{x}\in\mathbb{R}^{256\times 256\times1}$ and all of predicted water depth patches can be gathered to construct the final water depth map. The rainfall pattern vector $y\in\mathbb{R}^{1\times 12}$ is embedded into rainfall features by a rainfall mapping network, which are concatenated with terrain features in the channel dimension.

\begin{figure}[H]
\vspace{-0.5em}
\begin{minipage}{0.15\linewidth}
    \centering
    \includegraphics[width=1\linewidth]{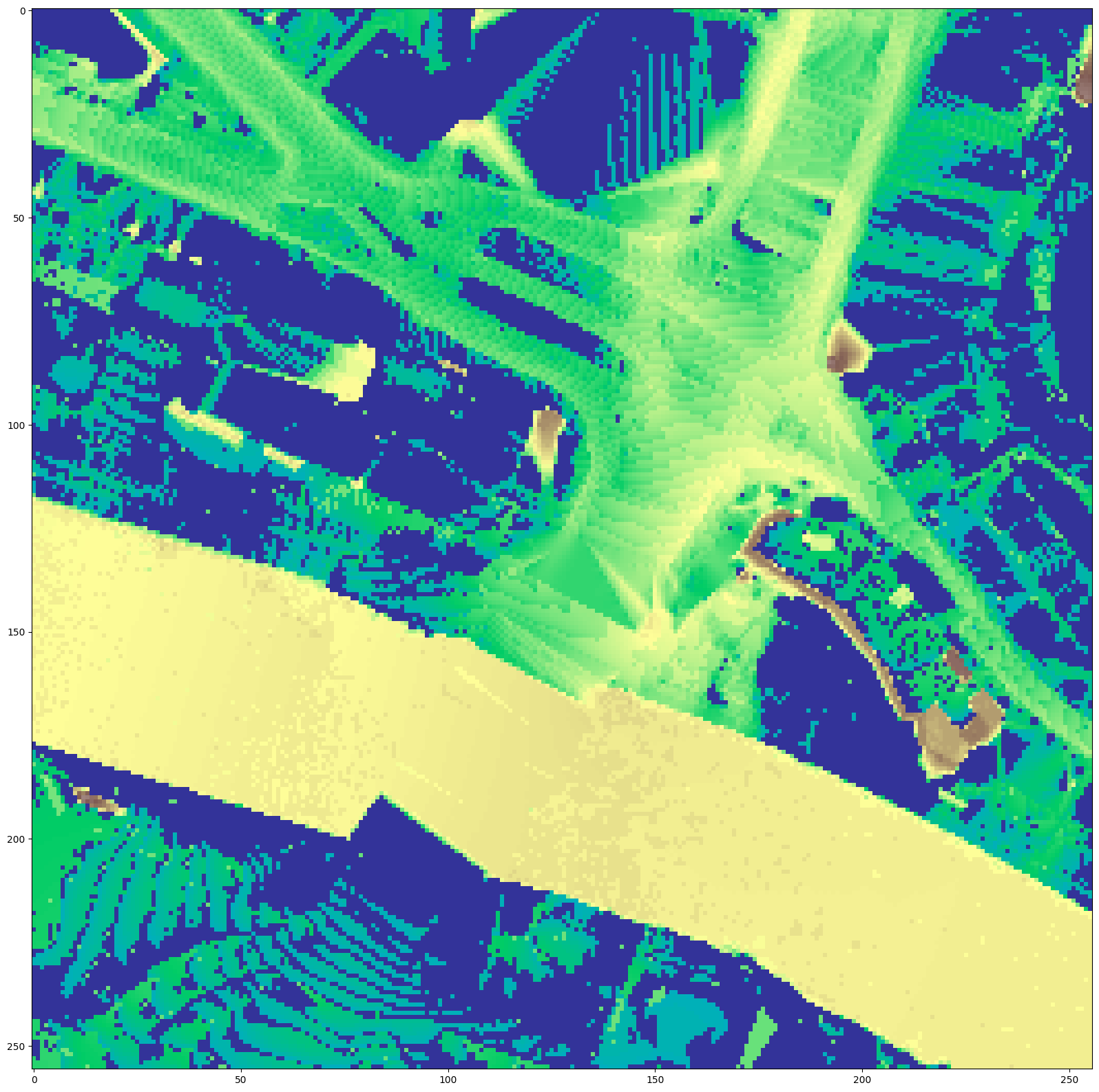}
    \centerline{\scriptsize \textbf{target}}
\end{minipage}
\hfill
\begin{minipage}{0.15\linewidth}
    \centering
    \includegraphics[width=1\linewidth]{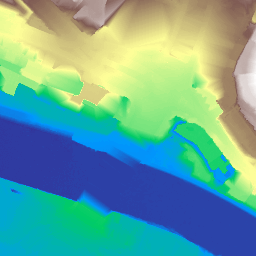}
    \centerline{\scriptsize (a) dem}
\end{minipage}
\hfill
\begin{minipage}{0.15\linewidth}
    \centering
    \includegraphics[width=1\linewidth]{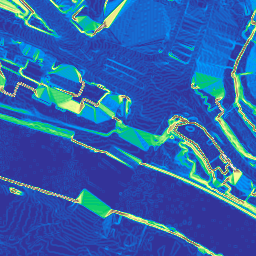}
    \centerline{\scriptsize (b) slope}
\end{minipage}
\hfill
\begin{minipage}{0.15\linewidth}
    \centering
    \includegraphics[width=1\linewidth]{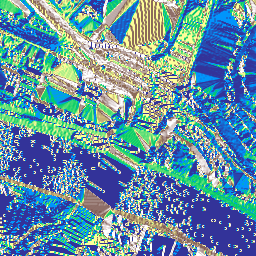}
    \centerline{\scriptsize (c) cosine}
\end{minipage}
\hfill
\begin{minipage}{0.15\linewidth}
    \centering
    \includegraphics[width=1\linewidth]{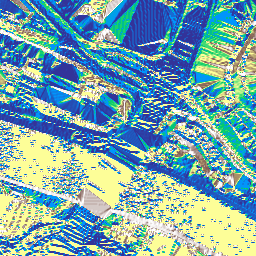}
    \centerline{\scriptsize (d) sine}
\end{minipage}
\hfill
\begin{minipage}{0.15\linewidth}
    \centering
    \includegraphics[width=1\linewidth]{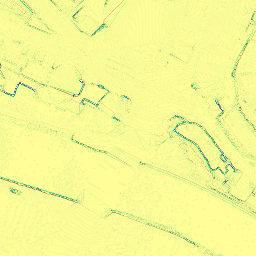}
    \centerline{\scriptsize (e) curvature}
\end{minipage}
\vspace{-0.5em}
\caption{Visualization for target water depth map $x$ and five channels of input terrain patch $d$. Here mask matrix is all 1 thus it is not be shown.}
\vspace{-0.5em}
\label{fig:image3}
\end{figure}

Fig~\ref{fig:image3} visualizes the input terrain patch which contains six channels including digital elevation model (DEM), mask matrix marking effective location (value of non-data areas is -1 and value of effective areas is 1), slope, cosine/sine (indicating aspect) and curvature. The target water depth patch $x$ and the input terrain patch $d$ share the consistent geographic contour.

\noindent\textbf{Adversarial loss.} 
The generator $G$ uses the given terrain patches $d$ and rainfall pattern vectors $y$ to predict water depth patch. 
Inspired by Pix2Pix \cite{IsolaZZE17}, our discriminator $D$ adopt PatchGAN loss, \ie,  the generated water depth patch is divided into 70$\times$70 smaller sub-patches and each sub-patch can get a probability score, respectively. 
These scores are averaged to vote for the final probability through $D$ to indicate whether the input water depth patch is realistic. 
Formally,
\begin{equation}
\begin{aligned}
 \mathcal{L}_{adv} =& \mathbb{E}_{{x} \sim p_{\text{data}}}\left[\log D_{src}\left(x\mid d\right)\right]+\\
 & \mathbb{E}_{{\hat{x}} \sim p_{g}}\left[\log\left(1-D_{src}\left(\hat{x}\mid d\right)\right)\right], 
\end{aligned}
\end{equation}
where $D_{src}$ denotes the vanilla branch of $D$ that has an adversarial convolutional head, $p_{\text{data}}$ and $p_g$ represent the distributions of ground-truth and the generated water depth data.

\noindent\textbf{Rainfall regression loss.} 
We hope that $D$ can extract original rainfall pattern features from real water depth maps as much as possible to help it distinguish fake ones from the input. 
Meanwhile, $G$ is intended to be sensitive to different rainfall patterns, which promotes the generation under different rainfall conditions. 
Thus, we propose rainfall regression loss with \eref{reg-real} for $D$ and \eref{reg-fake} for $G$, respectively,
\begin{equation}
    \mathcal{L}_{reg}^{r}=\mathbb{E}_{{x} \sim p_{\text{data}}}\left[\frac{1}{12} \sum_{i=0}^{11}\lvert D_{reg}\left(x\mid d\right)_{i}- y_{i}\lvert \right], 
    \label{reg-real}
\end{equation}
\begin{equation}
     \mathcal{L}_{reg}^{f}=\mathbb{E}_{\hat{x} \sim p_{g}}\left[\frac{1}{12} \sum_{i=0}^{11} \lvert D_{{reg}}\left(\hat{x}\mid d\right)_{i}- y_{i}\lvert \right], 
     \label{reg-fake}
\end{equation}
where $y_i$ is the $i$-th element of rainfall vector $y$.

\noindent\textbf{Reconstruction loss.} 
To generate high-quality water depth maps and reduce the prediction error, we use the reconstruction loss to optimize the generator: 
\begin{equation}
     \mathcal{L}_{rec}=\mathbb{E}_{{x} \sim p_{\text{data}}}\left[\| G\left(d,y\right)- x\|_{1} \right] 
     \label{prediction}
\end{equation}

\noindent\textbf{The overall objective.} 
The training objective for $D$ and $G$ are the weighted combination of aforementioned losses, \ie,
\begin{equation}
    \mathcal{L}_{D}=-\lambda_{adv}\mathcal{L}_{adv} + \lambda_{reg}\mathcal{L}_{reg}^{r}
    \label{5}
\end{equation}

\begin{equation}
     \mathcal{L}_{G}=\lambda_{adv}\mathcal{L}_{adv} + \lambda_{reg}\mathcal{L}_{reg}^{f}+\lambda_{rec}\mathcal{L}_{rec}
     \label{6}
\end{equation}

\subsection{Hierarchical Terrain Spatial Attention}
\begin{figure}[htbp]
\begin{minipage}{0.19\linewidth}
    \centering
    \includegraphics[width=1\linewidth]{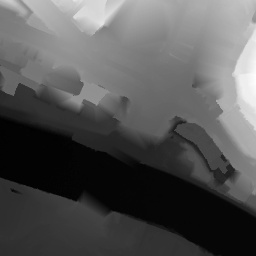}
    \centerline{\scriptsize Terrain patch}
\end{minipage}
\begin{minipage}{0.19\linewidth}
    \centering
    \includegraphics[width=1\linewidth]{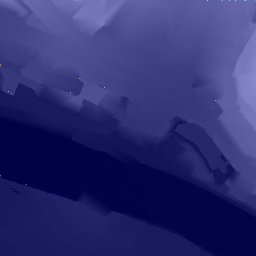}
    \centerline{\scriptsize HTA layer 4}
\end{minipage}
\hfill
\begin{minipage}{0.19\linewidth}
    \centering
    \includegraphics[width=1\linewidth]{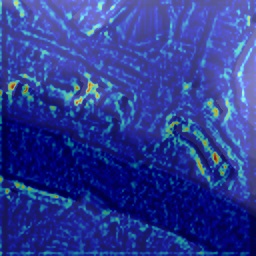}
    \centerline{\scriptsize HTA layer 3}
\end{minipage}
\begin{minipage}{0.19\linewidth}
    \centering
    \includegraphics[width=1\linewidth]{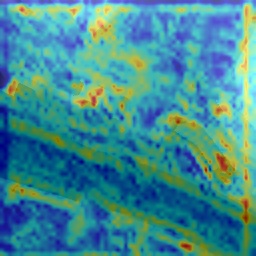}
    \centerline{\scriptsize HTA layer 2}
\end{minipage}
\hfill
\begin{minipage}{0.19\linewidth}
    \centering
    \includegraphics[width=1\linewidth]{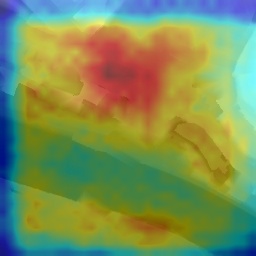}
    \centerline{\scriptsize HTA layer 1}
\end{minipage}
\vspace{-0.5em}
\caption{Grad-CAM heatmaps for visualizing HTA maps.}
\vspace{-0.5em}
\label{fig:image2}
\end{figure}

Empirically, the importance of different scaled terrain features $\mathbf{F}_{d} \in \mathbb{R}^{C\times H \times W}$ are likely to be spatial-varying, \ie, different spatial areas deserves different concerns given by the model. To this end, we introduce hierarchical spatial attention to terrain features with different scales \cite{woo2018cbam} and the HTA maps are used as attention weight matrices. It encourages the generator to concentrate more on the terrain feature areas that are more likely to be dominant for prediction. We visualize the HTA maps of four layers in Fig~\ref{fig:image2} and find that the attention area changes from local to global as the network goes deeper. 

Specifically, we first perform average pooling and maximum pooling operations on each terrain feature map along the channel dimension. Then the obtained two maps $\mathbf{F}_{d}^{\mathbf{avg}} \in \mathbb{R}^{1\times H \times W}$ and $\mathbf{F}_{d}^{\mathbf{max}} \in \mathbb{R}^{1\times H \times W}$ are concatenated on the channel dimension and transformed by a convolution layer, constructing the hierarchical terrain spatial attention maps $\mathbf{M}_{\mathbf{s}}(\mathbf{F}_{d}) \in \mathbb{R}^{1\times H \times W}$, formulated as, 
\begin{equation}
\begin{aligned}
\mathbf{M}_{\mathbf{s}}(\mathbf{F}_{d}) &=\sigma\left(f^{7 \times 7}([\operatorname{AvgPool}(\mathbf{F}_{d}) ; \operatorname{Max} \operatorname{Pool}(\mathbf{F}_{d})])\right) \\
&=\sigma\left(f^{7 \times 7}\left(\left[\mathbf{F}_{d}^{\mathbf{avg}} ; \mathbf{F}_{d}^{\mathbf{max}}\right]\right)\right),\\
\end{aligned}
\end{equation}
where $\sigma$ means the activation function and $f^{7 \times 7}$ denotes a convolution operation of the filter size of $7 \times 7$. Finally, we element-wisely multiply the feature maps by the computed spatial attention map, \ie, 
\begin{equation}
\begin{aligned}
\mathbf{{F}_{d}}^{\prime} &=\mathbf{M}_{\mathbf{s}}\left(\mathbf{F}_{d}\right) \otimes \mathbf{F}_{d},   
\end{aligned}    
\label{8}
\end{equation}
where $\otimes$ is element-wise multiplication. The attention results are concatenated with the output of deconvolution layers in each decoding block as the input of the following convolution.

\subsection{Multi-Scale Rainfall Embedding}
The well-embedded rainfall features $\mathbf{F}_{r}$ can greatly benefit water depth map prediction. Therefore, we propose multi-scale rainfall embedding to help the model understand the effect of different rainfall patterns.
Specifically, the rainfall pattern vector $y$ is initially embedded by a linear projection and subsequently converted into coarse-to-fine rainfall features with a learnable mapping network. 
Since water depth feature maps are continuously upsampled during decoding, the multi-scale rainfall features are concatenated with the output of decoder blocks in the channel dimension.
\begin{equation}
\begin{aligned}
\mathbf{{F}_{r}}^{\prime}&=\sigma\left(f_{t}^{2 \times 2}(\mathbf{F}_{r})\right) 
\end{aligned}
\end{equation}
where $\sigma$ represents the activation function and $f_{t}^{2 \times 2}$ denotes a deconvolution operation with the filter size of $2 \times 2$.

\section{experiment}
\label{sec:typestyle}
\subsection{Setup}
\noindent\textbf{Datasets} 
We use a public dataset \cite{Guo2020DatadrivenFE} provided by ETH Zurich  for two particular catchments of Zurich and Portugal for evaluation, including terrain images, 18 rainfall patterns and corresponding hydrodynamic simulation. 
The terrain images are randomly cropped into 10,000 patches with the resolution of 256$\times$256. 
The rainfall patterns are one-hour rainfall hyetographs which consist of a sampling of rainfall every five minutes.
We randomly select 12 rainfall patterns as training set and the remaining 6 patterns as test set.

\noindent\textbf{Implementation details}  
During training, we use Adam optimizer with a unified learning rate of 2e-4 for all models. The batchsize is set as 32 and the number of epochs is 80. 
For hyper-parameters, we set $\lambda_{adv}=0.001$, $\lambda_{reg}=0.005$ and $\lambda_{rec}=1$.
For quantitative evaluation of performance between our proposed framework and baselines on test set, we leverage four metrics that are widely adopted in flood prediction: mean absolute error (MAE), R$^{2}$ (Ratio of the prediction variance which is explicable according to the input), CSI$_{0.05m}$ (Binary comparison on pixel above threshold 0.05m), $A_{NN}/{A_{HD}}$ (Ratio of total area flooded $>$0.05m). 

\begin{figure}[t]
\begin{minipage}{0.32\linewidth}
    \centering
    \includegraphics[width=1\linewidth]{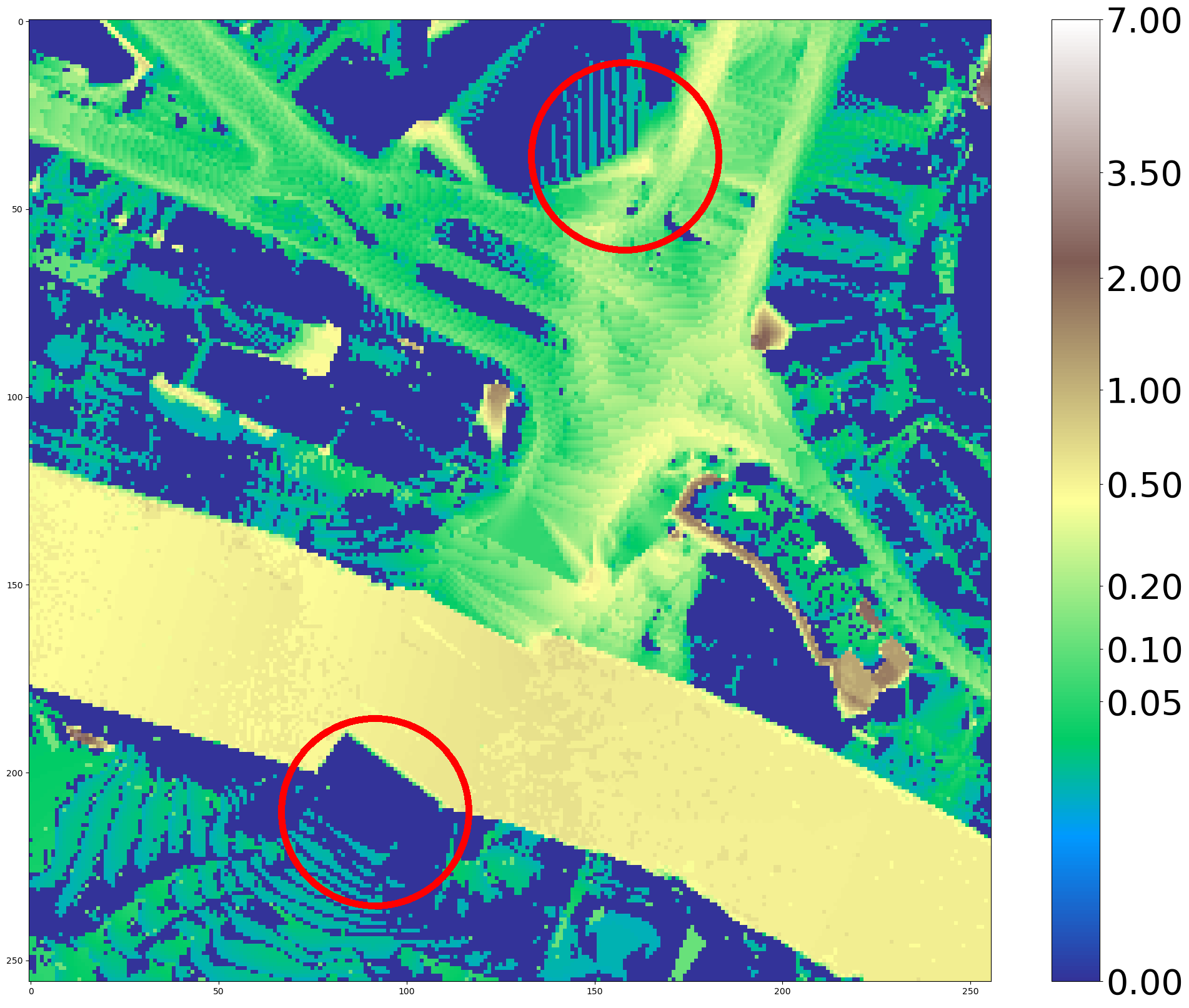}
\end{minipage}
\hfill
\begin{minipage}{0.32\linewidth}
    \centering
    \includegraphics[width=1\linewidth]{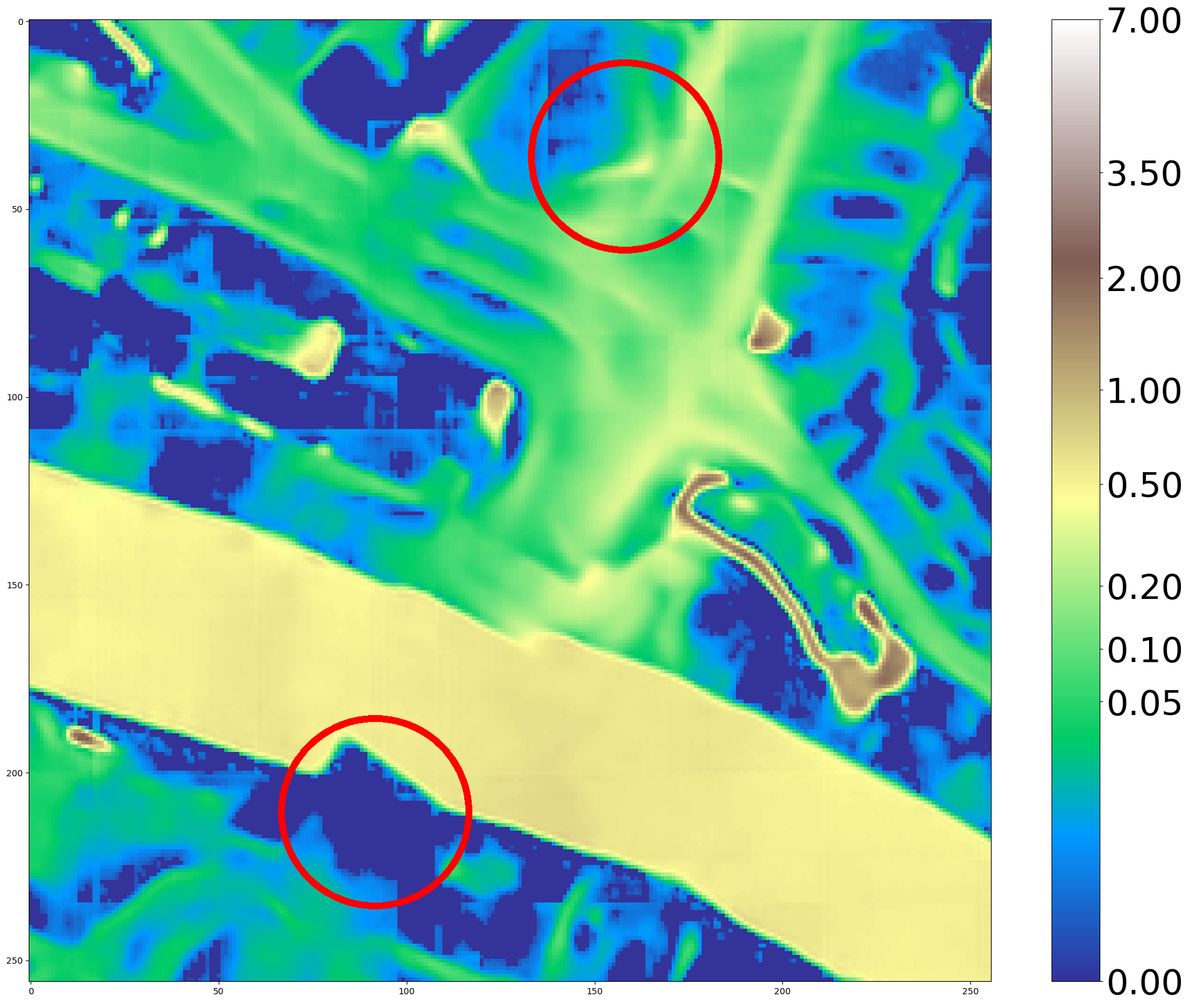}
\end{minipage}
\hfill
\begin{minipage}{0.32\linewidth}
    \centering
    \includegraphics[width=1\linewidth]{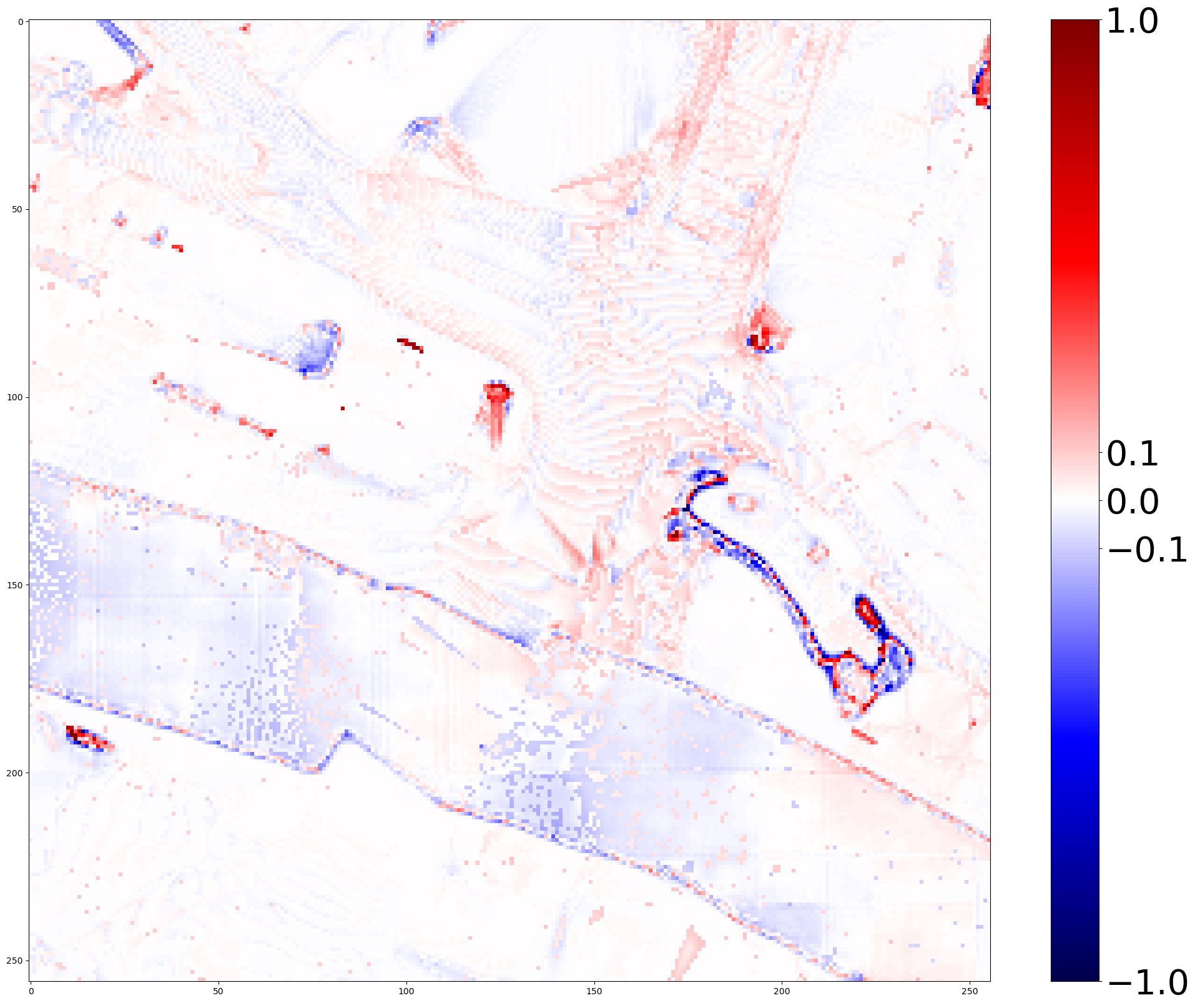}
\end{minipage}
\vfill
\begin{minipage}{0.32\linewidth}
    \centering
    \includegraphics[width=1\linewidth]{figures/tr100-2-truth-circle.png}
\end{minipage}
\hfill
\begin{minipage}{0.32\linewidth}
    \centering
    \includegraphics[width=1\linewidth]{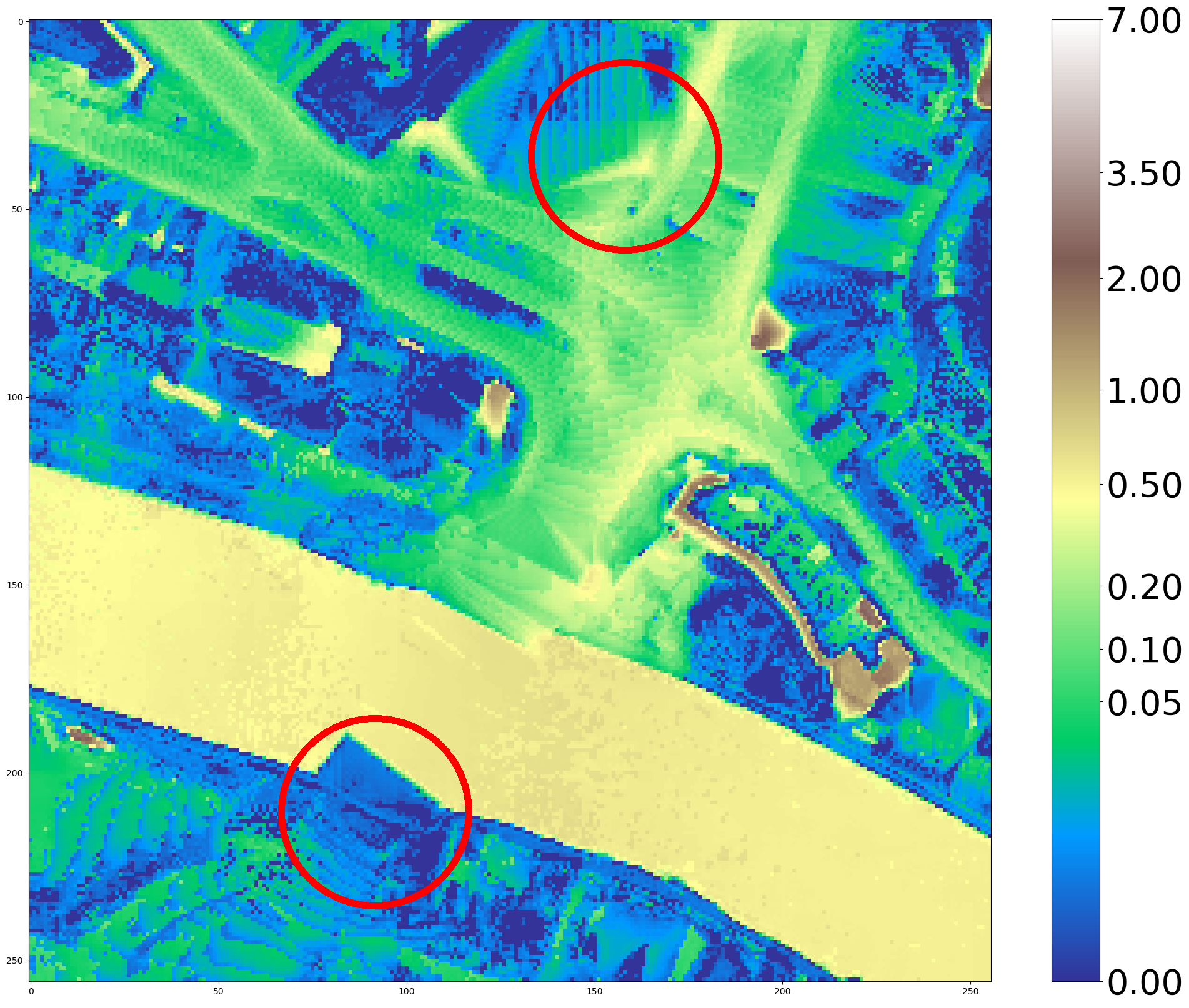}

\end{minipage}
\hfill
\begin{minipage}{0.32\linewidth}
    \centering
    \includegraphics[width=1\linewidth]{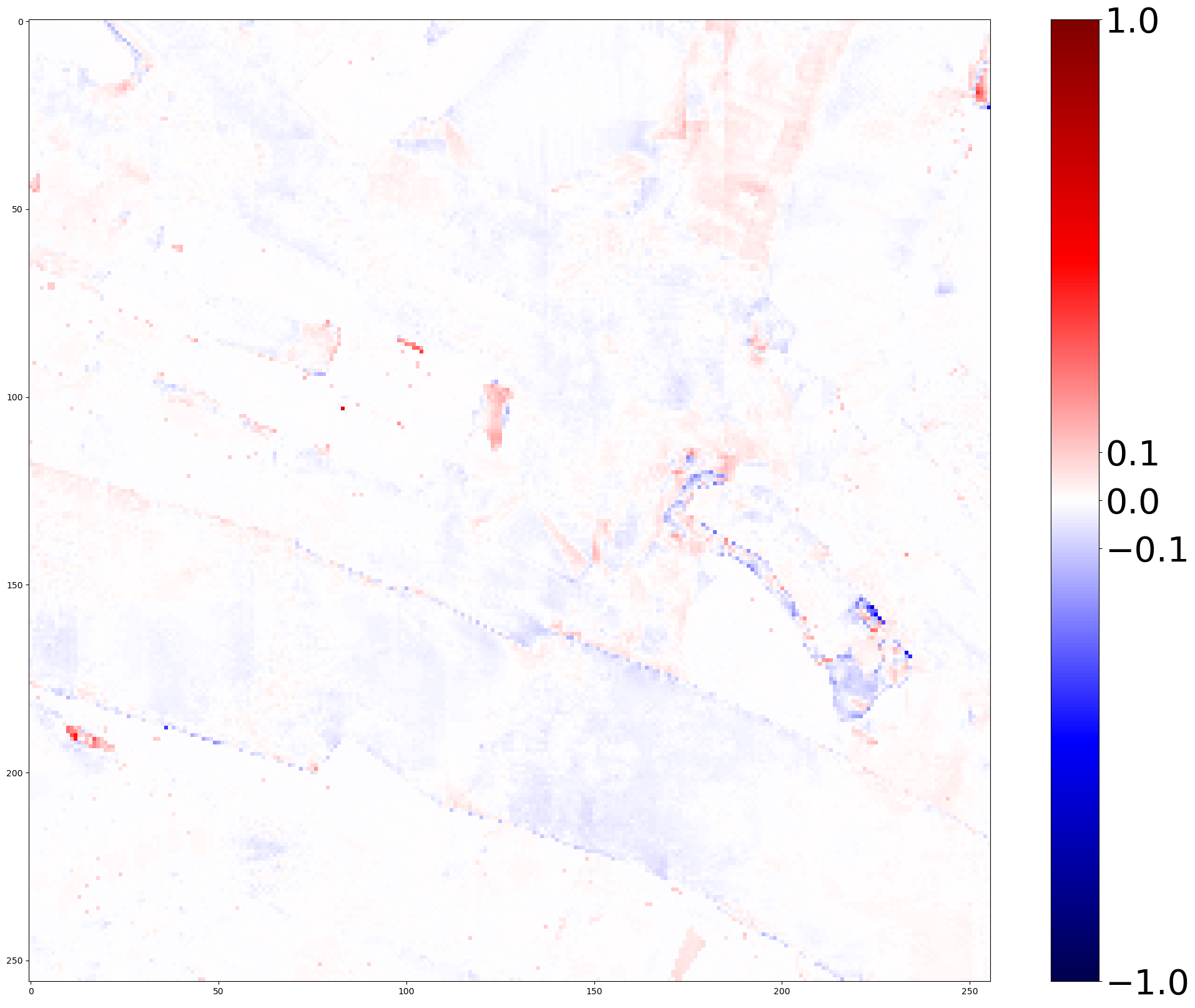}
\end{minipage}
\vfill
\begin{minipage}{0.32\linewidth}
    \centering
    \includegraphics[width=1\linewidth]{figures/tr100-2-truth-circle.png}
    \\ 
    (a) ground-truth
\end{minipage}
\hfill
\begin{minipage}{0.32\linewidth}
    \centering
    \includegraphics[width=1\linewidth]{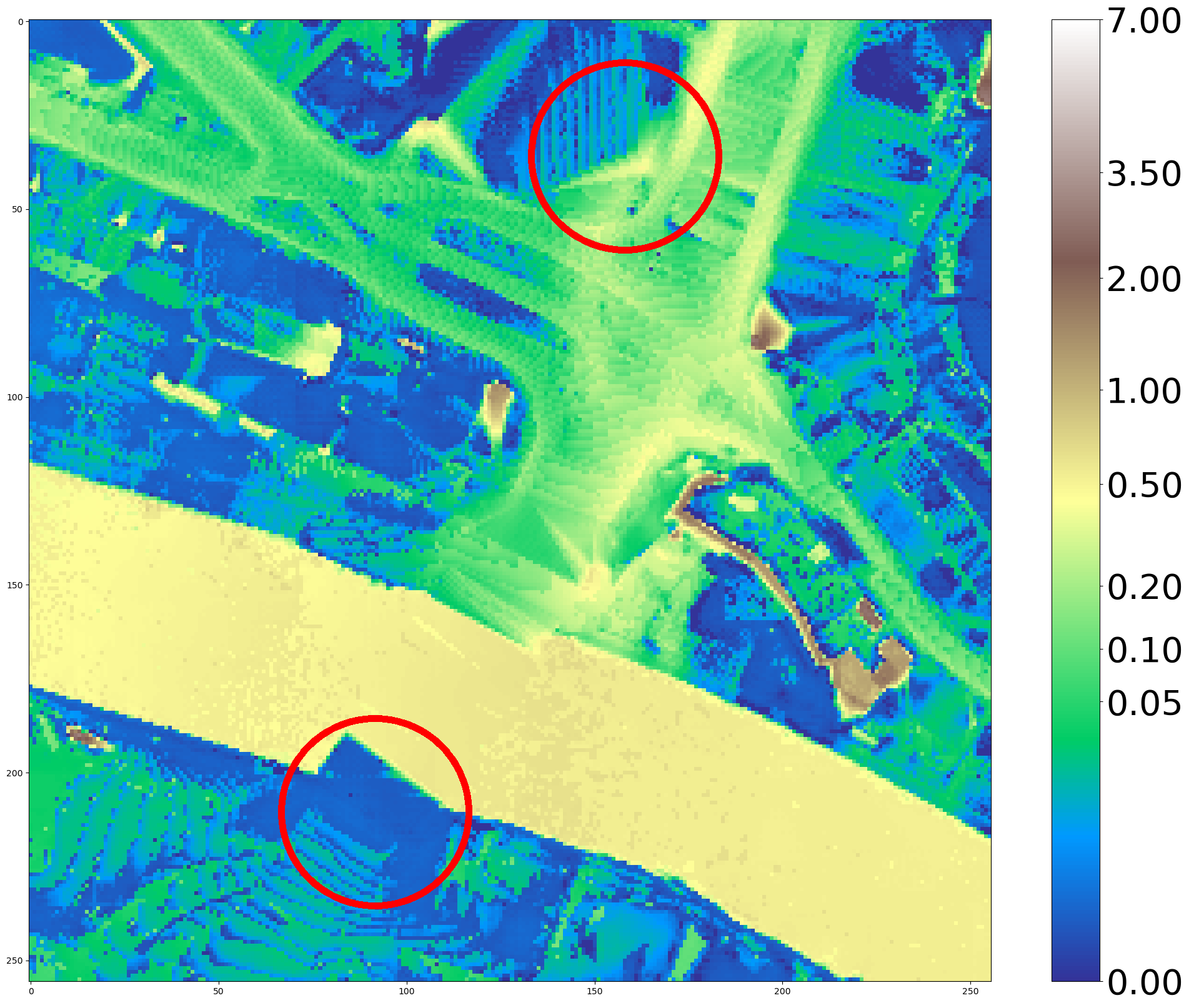}
    \\ 
    (b) predicted
\end{minipage}
\hfill
\begin{minipage}{0.32\linewidth}
    \centering
    \includegraphics[width=1\linewidth]{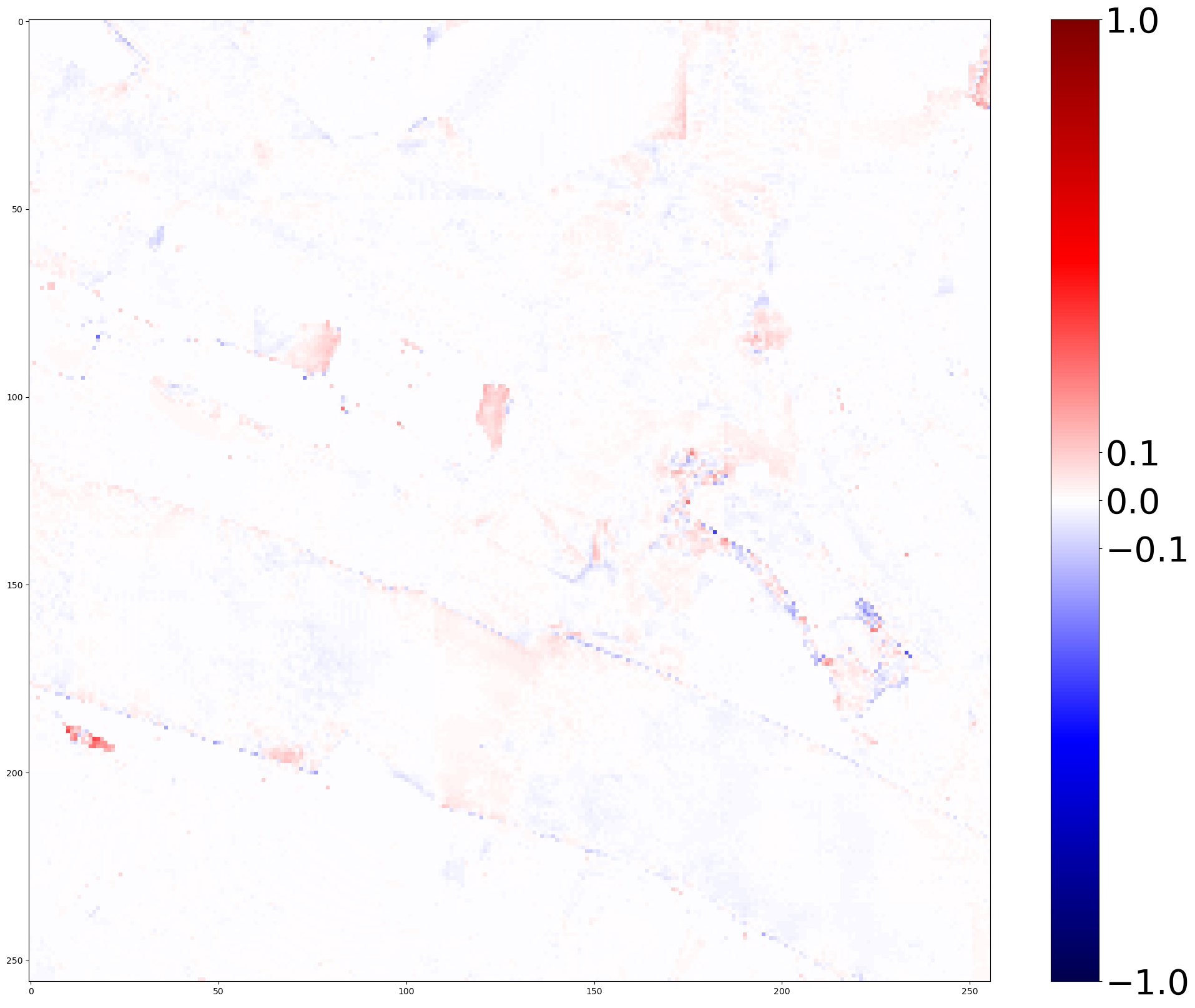}
    \\
    (c) error map
\end{minipage}
\caption{Visual comparison of Guo \etal\cite{Guo2020DatadrivenFE} (\textbf{top}), U-flood \cite{LOWE2021126898} (\textbf{middle}) and our method (\textbf{bottom}), including ground-truth, predicted flood images and error maps on the Zurich dataset.}
\label{fig:image1}
\end{figure}

\vspace{-1em}
\subsection{Comparison with baseline method}
We compare our work with the mainstream deep learning-based flood prediction methods and three popular image translation networks. Guo \etal \cite{Guo2020DatadrivenFE} and U-flood \cite{LOWE2021126898} use convolutional neural network (CNN) and its variants \cite{RonnebergerFB15} to predict water depth with terrain and rainfall as input. For image translation networks that have performed well in their expertise field, we transfer them to flood prediction as important baselines. Pix2Pix \cite{IsolaZZE17} incorporates the GAN supervision with end-to-end reconstruction to convert images from the original domain to the target domain. Based on Pix2Pix, GcGAN \cite{DBLP:conf/cvpr/FuGWB0T19} introduces geometric consistency to ensure that the manipulated input can produce results close to the ground-truth after the same geometric operation. TransUNet \cite{chen2021TransUNet} adopts the basic framework of U-Net and introduces vision transformer after the last encoder block to focus on the long-range relationship in the spatial space. 
As the results shown in \Tref{table1} and \Fref{fig:image1}, our method achieves much lower prediction error that greatly surpasses the baselines. Compared with other methods, our proposed framework shows strong generalization ability when handling different catchment datasets. 
For training time, our method takes only $\sim$2h 53min on a single RTX 3090Ti GPU and can be adapted into unseen flood conditions, while the hydrodynamic simulation needs to take $\sim$4h 54min \cite{Guo2020DatadrivenFE} for each flood condition.

\begin{table}[t]
    \scriptsize
    \centering
    \caption{Comparison of the performance of our proposed framework with mainstream deep flood prediction methods and image translation networks. The performance is better when $A_{NN}/{A_{HD}}$ is closer to 1.}
    \vspace{-1em}
    \setlength{\tabcolsep}{1mm}{
    \begin{tabular}{l|p{10mm}<{\centering}p{10mm}<{\centering}p{11mm}<{\centering}p{14mm}<{\centering}|c}
    \toprule
    \multirow{1}{*}{Method}
    & MAE$\downarrow$
    & R$^{2}\uparrow$
    & CSI$_{0.05m}$$\uparrow$
    & $A_{NN}/{A_{HD}}$
    \\
    \hline
    Guo \etal \cite{Guo2020DatadrivenFE}  & 153.2 & 64.3$\%$ & 0.653 & 90 $\%$ & \multirow{6}{*}{\rotatebox{90}{Zurich}}
    \\
    U-flood \cite{LOWE2021126898} & 
    64.2 & 84.9$\%$ & 0.853 & 98$\%$&
    \\
    Pix2Pix \cite{IsolaZZE17} & 
    74.1 & 82.6$\%$ & 0.841& 99$\%$ &
    \\
    GcGAN \cite{DBLP:conf/cvpr/FuGWB0T19} & 75.3 & 82.3$\%$ & 0.834 & 98$\%$ &
    \\
    TransUNet \cite{chen2021TransUNet} &
    89.5 & 79.0$\%$ & 0.805 & 97$\%$ &
    \\
    \cellcolor{gray!20}{\textbf{Ours}} & \cellcolor{gray!20}{\textbf{46.5}} & \cellcolor{gray!20}{\textbf{89.1$\%$}} &
    \cellcolor{gray!20}{\textbf{0.906}} &
    \cellcolor{gray!20}{\textbf{101$\%$}} & 
    \\
    \hline
    Guo \etal \cite{Guo2020DatadrivenFE} & 124.6 & 47.4$\%$ & 0.653 & 87$\%$ & \multirow{6}{*}{\rotatebox{90}{Portugal}}
    \\
   U-flood \cite{LOWE2021126898} & 68.2 & 71.9$\%$ & 0.683 & 76$\%$ &
    \\
    Pix2Pix  \cite{IsolaZZE17} & 135.0 & 43.4$\%$ & 0.375 & 46$\%$ &
    \\
    GcGAN \cite{DBLP:conf/cvpr/FuGWB0T19} & 124.9 & 47.8$\%$ & 0.412 & 46$\%$ &
    \\
    TransUNet  \cite{chen2021TransUNet} & 51.6 & 78.5$\%$ & 0.767 & 86$\%$ &
    \\
    \cellcolor{gray!20}{\textbf{Ours}} & \cellcolor{gray!20}{\textbf{34.2}} & \cellcolor{gray!20}{\textbf{85.7$\%$}} &
    \cellcolor{gray!20}{\textbf{0.850}} &
    \cellcolor{gray!20}{\textbf{97$\%$}} & 
    \\
    \bottomrule
    \end{tabular}
    }
    \label{table1}
\end{table}


\vspace{-1em}
\begin{table}[t]
	\footnotesize
    \centering
    \caption{Ablation results for each proposed component in our framework. ``HTA'' denotes hierarchical terrain spatial attention and ``MRE'' denotes multi-scale rainfall embedding.}
    \label{tab:table2}
    \setlength{\tabcolsep}{1mm}{
    \begin{tabular}{lp{6mm}<{\centering}p{7mm}<{\centering}p{11mm}<{\centering}p{13mm}<{\centering}}
    \toprule
    \multirow{1}{*}{Ablation Setting}
    & MAE$\downarrow$
    & R$^{2}\uparrow$
    & CSI$_{0.05m}$$\uparrow$
    & $A_{NN}/A_{HD}$
    \\
    \midrule
    Base & 88.1 & 79.3$\%$ & 0.800 & 96$\%$
    \\
    + HTA  &  60.4 & 85.8$\%$ & 0.868 & 98$\%$
    \\
    + MRE & 49.3 & 88.4$\%$  & 0.892 & 99$\%$
    \\
    + HTA + MRE & 48.4 & 88.6$\%$ &  0.895 & 100$\%$
    \\
    + HTA + MRE + $\mathcal{L}_{gan}$ & 47.0 & 88.9$\%$ &  0.904 & 101$\%$
    \\
    \cellcolor{gray!20}{+ HTA + MRE + $\mathcal{L}_{gan}$ + $\mathcal{L}_{reg}$} & \cellcolor{gray!20}{46.5} &
    \cellcolor{gray!20}{89.1$\%$} &
    \cellcolor{gray!20}{0.906} &
    \cellcolor{gray!20}{101$\%$} 
    \\
    \bottomrule
    \end{tabular}
    }
\label{table2}
\end{table}

\subsection{Ablation study}
We ablate each of our designed components to demonstrate their significance in \Tref{table2}. 
It is worth noting that HTA and MRE contribute the most significant improvements. 
It verifies the validity of our intuition on fully utilizing the terrain characteristics and rainfall information.

\section{conclusion}
In this paper, we find that previous flood prediction methods fail to make sufficient use of terrain and rainfall information, which greatly diminishes the performance. To alleviate the impacts, we propose a novel framework which contains two enhancement components, \ie, hierarchical terrain spatial attention and multi-scale rainfall embedding. The experiments show that our method can significantly improve the prediction accuracy while maintaining the timeliness.

\newpage
\bibliographystyle{IEEEbib}
\bibliography{strings,refs}

\end{document}